\documentclass[10pt,twocolumn]{article} 
\usepackage{simpleConference}
\usepackage{times}
\usepackage{graphicx}
\usepackage{amssymb}
\usepackage{url,hyperref}
\usepackage[nocompress]{cite}
\usepackage{amsmath}
\usepackage{amsfonts}
\usepackage{algorithm}
\usepackage{algorithmic}
\usepackage{array}
\usepackage[caption=false,font=footnotesize,labelfont=sf,textfont=sf]{subfig}
\usepackage[caption=false, font=footnotesize]{subfig}
\usepackage{caption}
\usepackage{graphicx}
\usepackage[]{footmisc}
\usepackage{hyperref}
\usepackage{lipsum}
\usepackage[dvipsnames]{xcolor}

\begin{document}

\title{Recurrent Auto-Encoder with Multi-Resolution Ensemble and Predictive Coding for Multivariate Time-Series Anomaly Detection}

\author{Heejeong~Choi, Subin~Kim, and~Pilsung~Kang* \\
\\
The School of Industrial \& Management Engineering, Korea University \\
\{heejeong\_choi, subin-kim, pilsung\_kang\}@korea.ac.kr  \\
}

\maketitle
\thispagestyle{empty}

\begin{abstract}
As large-scale time-series data can easily be found in real-world applications, multivariate time-series anomaly detection has played an essential role in diverse industries. It enables productivity improvement and maintenance cost reduction by preventing malfunctions and detecting anomalies based on time-series data. However, multivariate time-series anomaly detection is challenging because real-world time-series data exhibit complex temporal dependencies. For this task, it is crucial to learn a rich representation that effectively contains the nonlinear temporal dynamics of normal behavior. In this study, we propose an unsupervised multivariate time-series anomaly detection model named RAE-MEPC which learns informative normal representations based on multi-resolution ensemble and predictive coding. We introduce multi-resolution ensemble encoding to capture the multi-scale dependency from the input time series. The encoder hierarchically aggregates the temporal features extracted from the sub-encoders with different encoding lengths. From these encoded features, the reconstruction decoder reconstructs the input time series based on multi-resolution ensemble decoding where lower-resolution information helps to decode sub-decoders with higher-resolution outputs. Predictive coding is further introduced to encourage the model to learn the temporal dependencies of the time series. Experiments on real-world benchmark datasets show that the proposed model outperforms the benchmark models for multivariate time-series anomaly detection.
\end{abstract}

\section{Introduction}
A significant amount of multivariate time-series data has been accumulated in complex systems such as smart factories, power plants, and cyber-security \cite{lee2018machine, laubscher2019time, pokhrel2017cybersecurity}. To identify potential threats, it is critical to monitor the operating conditions of these systems \cite{wang2020detecting}. In many application domains, time-series anomaly detection is a vital component of operating these monitoring systems, as failure to take proper action in response to an early warning sign might cause an actual accident, resulting in substantial loss \cite{wu2016survey}. Anomalies are defined as data points that deviate significantly from normal observations \cite{chandola2009anomaly}. The objective of time-series anomaly detection is to identify anomalies in each time step of a specified length of time-series data \cite{chalapathy2019deep}. Therefore, many industrial systems that regularly generate data require an accurate time-series anomaly detection model to improve system productivity.

An effective time-series anomaly detection method should sufficiently capture the nonlinear temporal information of the normal time-series data while also detecting previously unknown anomalies \cite{shen2020timeseries}. However, this task is extremely difficult. First, anomalies are difficult to detect in practice \cite{canizo2019multi}. Therefore, time-series anomaly detection is usually implemented unsupervised, with only normal samples. Second, multivariate time-series data exhibit complex temporal dependencies and stochasticity \cite{su2019robust}. Most existing methods address this issue by focusing on learning good representations that capture the long-term dependency of the normal patterns.

Time-series anomaly detection based on deep learning has been studied in two main categories: 1) prediction-based methods and 2) reconstruction-based methods \cite{geiger2020tadgan}. First, prediction-based methods learn the features of the normal time-series data by predicting the values of future time steps based on the previous time-series data \cite{malhotra2015long, hundman2018detecting, shalyga2018anomaly, kravchik2018detecting, lai2018modeling}. These methods assume that a predictive model cannot accurately estimate abnormal patterns when trained on normal data only. During inference, they identify specific time segments as abnormal if the difference between the actual and predicted values for the segment is above a predefined threshold. Most prediction-based methods are based on long short-term memory (LSTM) \cite{hochreiter1997long} and gated recurrent units \cite{cho2014learning} that can model temporal features. 

Second, reconstruction-based methods learn normal patterns by encoding normal time-series data into compressed latent vectors and then reconstructing (decoding) the input from the latent vectors \cite{malhotra2016lstm, park2018multimodal, zong2018deep, yoo2019recurrent, yoo2019recurrent, kieu2019outlier, shen2021time}. These approaches assume that anomalies cannot be reconstructed well by the model trained only with normal patterns because mapping unseen anomalies into the latent space would result in a large reconstruction loss. Most reconstruction-based methods are based on an auto-encoder, which consists of an encoder for capturing the temporal features of the input time-series data and a decoder to reconstruct the input from the latent encoded vector. In these methods, the time segments are detected as anomalies if the difference between the input and reconstructed outputs is large. Moreover, generative adversarial networks (GANs) \cite{goodfellow2014generative} have been used in reconstruction-based methods to regularize reconstruction errors \cite{li2019mad, bashar2020tanogan, geiger2020tadgan, liu2022time}. These methods enable the model to learn the probability distribution of normal data by reconstructing samples via an adversarial process between the generator and discriminator. In this framework, a generator learns the data distribution and the discriminator estimates the probability that a given instance actually comes from the training data rather than the synthetic data produced by the generator. Reconstruction-based methods using a GAN assume that a generator with a normal distribution cannot adequately reconstruct samples with abnormal patterns. Based on this assumption, instances with a large probability difference between actual and reconstructed data are detected as anomalies. One critical disadvantage of reconstruction-based GAN methods is that their training processes are often unstable because of mode collapse and nonconvergence issues \cite{audibert2020usad}.

In this study, we propose the recurrent auto-encoder with multi-resolution ensemble and predictive coding (RAE-MEPC) to extract informative time-series representations by combining the advantages of reconstruction- and prediction-based methods. The proposed RAE-MEPC captures temporal information at different resolution levels in both the encoding and decoding stages. In particular, the encoder consists of multiple sub-encoders with different encoding lengths. Each sub-encoder extracts the temporal features on a different scale. A sub-encoder with a short encoding length can focus on global patterns, but a sub-encoder with a long encoding length can capture more local characteristics. Subsequently, their features are hierarchically consolidated into a final time-series representation with multi-scale dependency. Based on multi-resolution ensemble decoding, the reconstruction decoder of RAE-MEPC reconstructs the input time series from the encoded features. The output from the sub-decoder with the highest resolution is used as the ensemble output. Furthermore, the prediction-based method was employed as an auxiliary task to extract temporal information from the perspective of a prediction task. Adding the prediction decoder to the reconstruction decoder uses past latent features to predict future time series. Once the model is trained on a normal time series, RAE-MEPC can detect anomalies based on the difference between the input time series and the reconstructed output.

The main contributions of this study can be summarized as follows:
\begin{description}
    \item[$\bullet$] We propose a multi-resolution ensemble encoding method to learn the multi-scale dependency of time-series data.
    \item[$\bullet$] We introduce predictive coding to capture the temporal information of time-series data from the perspective of prediction and reconstruction tasks.
    \item[$\bullet$] Based on the experimental results for both univariate and multivariate time-series datasets, our proposed method outperforms well-known benchmark models.
\end{description}

The remainder of this paper is organized as follows. In Section 2, we briefly review reconstruction-based time-series anomaly detection and methods for improving the quality of representation. Section 3 describes our proposed method RAE-MEPC with multi-resolution ensemble reconstruction and predictive coding. In Section 4, the experimental settings are explained, followed by the experimental results. Finally, we summarize our study in Section 5.

\section{Related Work}
\subsection{Time-Series Anomaly Detection}
\subsubsection{Prediction-Based Methods}
Prediction-based methods capture information from normal time-series data by predicting the values of future time sequences from given time-series data. Most approaches employ a recurrent neural network (RNN) as the main building block because it is known to be useful for learning long-term relationships. LSTM-AD \cite{malhotra2015long} was proposed to detect anomalies with higher-level temporal features. The architecture of LSTM-AD is a stacked LSTM that predicts the time series over several time steps. In this method, anomalies can be detected based on prediction errors modeled as a multivariate Gaussian distribution. In \cite{hundman2018detecting}, an LSTM-based time-series anomaly detection model that learns normal patterns in prediction tasks was proposed. A complementary unsupervised and nonparametric anomaly thresholding approach, which automatically determines thresholds for data streams characterized by varying behavior and value ranges, was also proposed to address diversity, nonstationarity, and noise issues. In \cite{shalyga2018anomaly}, genetic algorithms were used to determine the best prediction-based time-series anomaly detection architecture. In this model, exponentially weighted smoothing, mean $p$-powered error measure, individual error weights for each variable, and disjoint prediction windows are introduced to improve anomaly detection performance. In \cite{kravchik2018detecting}, a unified prediction-based model was proposed, including an RNN and a convolutional neural network (CNN). This model detects anomalies based on statistical deviation of the predicted values from the observed values. LSTNet \cite{lai2018modeling} was proposed to model a mixture of long- and short-term patterns in time-series data. This method also uses an RNN and a CNN to extract diverse time-series trends. It also introduces the classic autoregressive model as a linear component to address the scale-insensitive problem of the neural network model.

\subsubsection{Reconstruction-Based Methods}
Reconstruction-based methods learn normal patterns in the process of reconstructing normal time series. Most approaches used recurrent auto-encoder (RAE) as a backbone model where both encoder and decoder are based on the RNN. EncDec-AD \cite{malhotra2016lstm} was proposed to model temporal information effectively. This LSTM auto-encoder is trained to reconstruct normal data and then detect anomalies based on the likelihood of a reconstruction error. LSTM-VAE \cite{park2018multimodal} was proposed to address the issue that the fusion of high-dimensional and heterogeneous modalities is complex in model-based anomaly detection. This model introduces a progress-based varying prior to fusing the signals and reconstructing their distribution. It uses a reconstruction-based anomaly score and detects data with a score higher than the state-based threshold as anomalies. DAGMM \cite{zong2018deep} was proposed to preserve important information of a normal time series in a low-dimensional space. This model is a decoupled architecture of an auto-encoder and Gaussian mixture model. It uses an auto-encoder to generate a low-dimensional representation, which is then fed into a Gaussian mixture model. These two components in DAGMM were simultaneously optimized for consistent optimization goals. This joint optimization aims to balance the reconstruction of the auto-encoder and density estimation of the latent representation. A recurrent reconstructive network \cite{yoo2019recurrent} was proposed to overcome the limitations of RAE in handling only fixed-length inputs. This method includes self-attention, hidden state forcing, and skip transitions. With the self-attention mechanism and hidden state forcing, the encoder and decoder can effectively manage input sequences of varying lengths. A skip transition is introduced to improve the reconstruction performance. The RAE-ensemble \cite{kieu2019outlier} was proposed to avoid overfitting of the RAE. The two solutions were built based on the sparsely connected RNN, which can produce multiple auto-encoders with different structures. In the RAE-ensemble, multiple auto-encoders are combined in both the independent and shared frameworks. These frameworks can prevent some auto-encoders from being overfitted by introducing an ensemble. RAMED \cite{shen2021time} proposed to alleviate the error accumulation problem caused in the decoding step of RAE. This model introduced multi-resolution ensemble decoding to the RAE-ensemble. It shares the information between multiple decoders with different decoding lengths by using lower-resolution information for higher-resolution decoding. A multi-resolution shape-forcing loss was further introduced to encourage the reconstructed outputs at multiple resolutions to match the global characteristics of the input time series. Moreover, reconstruction-based methods with adversarial learning learn the normal distribution by reconstructing samples based on GAN architecture. MAD-GAN \cite{li2019mad} was proposed to exploit the spatio-temporal correlation and dependencies of the entire variable. This model, with the generator and discriminator, detects anomalies using a novel anomaly score that is relevant to discrimination and reconstruction. TAnoGAN \cite{bashar2020tanogan} was proposed to detect anomalies in small amounts of data. This unsupervised method introduces adversarial training in LSTM to decrease the model parameters. TadGAN \cite{geiger2020tadgan} was proposed to address scalability and portability issues in time-series anomaly detection. This method uses the LSTM architecture for the generator and discriminator to capture the temporal correlations of time-series distributions. TadGAN is trained with cycle consistency loss to effectively capture normal patterns. It detects anomalies based on novel anomaly scores to combine the reconstruction errors and loss from the discriminator. BeatGAN \cite{liu2022time} was proposed to detect anomalies based on normal data. It comprises a one-dimensional CNN and RNN, which are trained adversarially. Data augmentation with dynamic time warping (DTW) is added for regularization and robustness.

%Figure 1
\begin{figure*}[!t]
    \centering
    \includegraphics[width=1\textwidth]{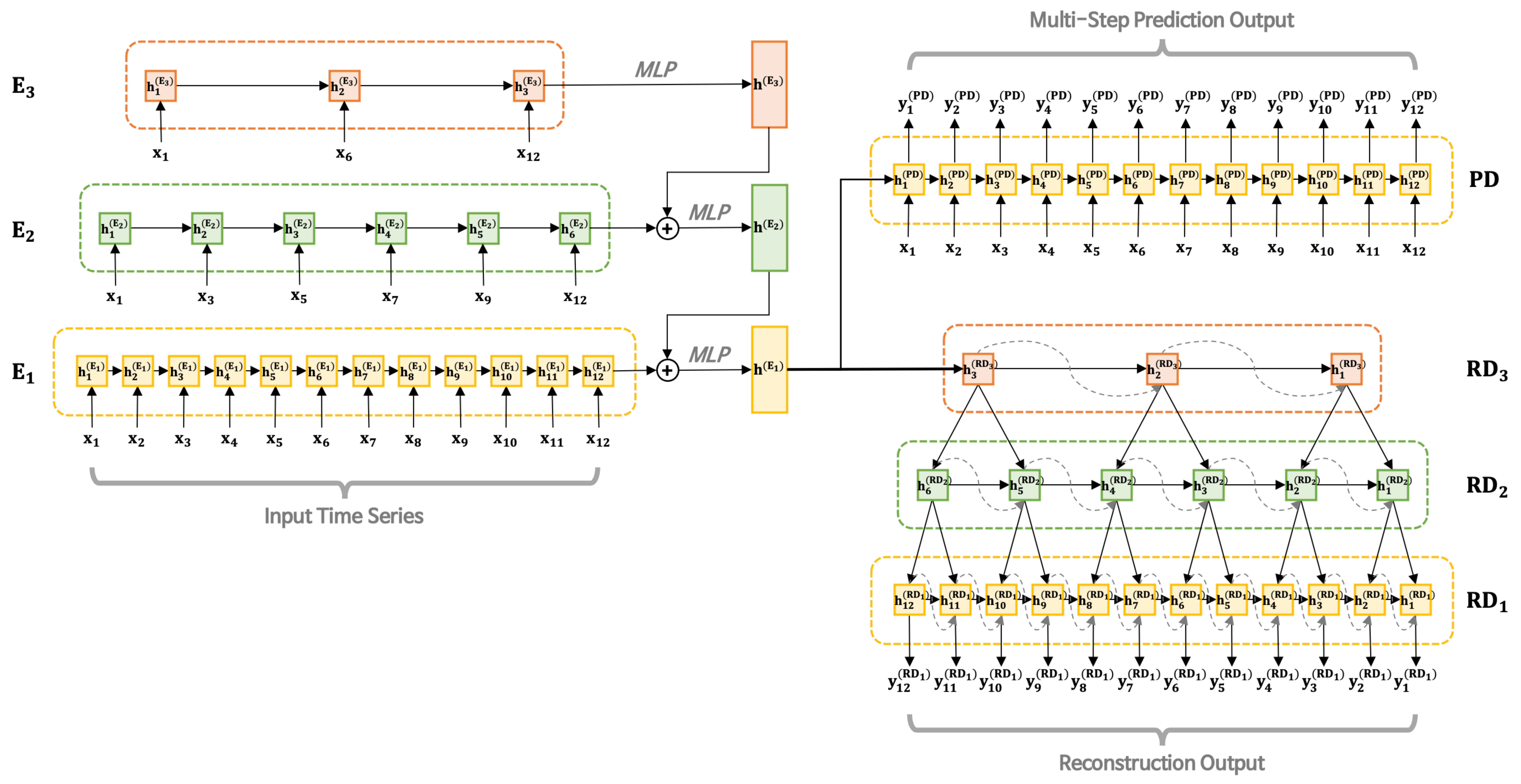}
    \caption{Architecture of the proposed method RAE-MEPC.}
    \label{fig1}
\end{figure*}

\subsection{Modeling of Long-Term and Multi-Scale Dependency}
Various methods have been researched to extract long-term and multi-scale dependency from sequential time-series data. Most approaches are based on the RNN, which is a representative model for temporal features. A hierarchical RNN \cite{el1996hierarchical} was proposed to improve the long-term dependency of RNN. In this model, domain-specific priori knowledge is introduced to give meaning to the hidden variables representing the past context. This model hierarchically integrates temporal information of diverse delays and resolutions. A hierarchical multi-scale RNN \cite{chung2016hierarchical} was proposed to resolve the issue that learning both hierarchical and temporal representation is difficult in RNN. This method consists of multiple recurrent layers with different time scales. It can extract hierarchical latent features with various temporal dependencies in a sequence by sharing the information of each layer. A temporal pyramid RNN \cite{ma2020temporal} was proposed to learn long-term and multi-scale dependencies in sequential data. The architecture is built by stacking multiple recurrent layers of sub-pyramids. In this method, the input sequence of the higher layer is a large-scale aggregated state sequence produced by the sub-pyramids in the previous layer. It can explicitly learn multi-scale dependencies using multi-scale input sequences of different layers. Furthermore, a shortcut path is added to the output of each sub-pyramid to shorten the gradient feedback path of each layer and avoid vanishing gradient problems in RNNs.

\subsection{Representation Learning with Predictive Coding}
Various methods have been studied for learning temporal features in diverse domains relevant to time series. In video representation learning, various methods have been proposed for learning temporal embedding based on the future frame prediction. \cite{srivastava2015unsupervised} improved the video representation based on the reconstruction and prediction task. This method consists of one LSTM encoder and two LSTM decoders. The encoder maps the input video sequence into fixed-length representations, while two decoders reconstruct the input sequence and predict future sequences. In \cite{mathieu2015deep}, a model based on a CNN that generates a future frame from an input video sequence was proposed. This method is based on a multi-scale architecture in which an adversarial training method and an image gradient difference loss function are applied to address the blurry predictions obtained from the mean squared error. Inspired by predictive coding, a predictive neural network \cite{lotter2016deep} was proposed to overcome the problem that unsupervised learning leverages unlabeled examples to learn about the structure of a domain. This model can learn video representations in predicting future frames from a video sequence. In this model, each layer predicts future frames locally, before passing them on to the next layer, which then predicts the entire frame. A dense predictive coding \cite{han2019video} was proposed to learn a spatio-temporal embedding from a video in a self-supervised way. This method can encode dense sequences based on spatio-temporal blocks, which sequentially predict future representations. Furthermore, a curriculum training scheme was proposed to learn semantic representation by encoding only spatial-temporal signals with slow changing. In this training, the future representation is predicted gradually with less temporal context.

\section{Proposed Method}
\subsection{Recurrent Auto-Encoder with Multi-Resolution Ensemble and Predictive Coding}
In this paper, we propose multivariate time-series anomaly detection model called RAE-MEPC, which can model multi-scale and temporal dependency in time series using multi-resolution ensemble and predictive coding. Our proposed method aims to learn the normal patterns by reconstructing the input time series and predicting future time series based on the informative encoded features. Let $\mathrm{X}=[\mathrm{x}_1, \mathrm{x}_2, \dotsm , \mathrm{x}_T] \in \mathbb{R}^{d \times T}$ be the input time window, which is $T$ time steps in time-series length. Each $t$-th time step in input time window $\mathrm{x}_t=[x_1, x_2, \dotsm , x_d] \in \mathbb{R}^{d}$ has $d$ number of variables. In this study, we detect anomalies in an unsupervised manner based on the assumption that most multivariate time series in the training data are normal. The proposed RAE-MEPC comprises an encoder and two decoders. The encoder maps the input time series into compressed representations with multi-scale dependency, while two decoders are built to reconstruct the input time series and predict future time steps. Fig. \ref{fig1} shows the architecture of our proposed model, which learns normal patterns based on the three components as follows: 1) multi-resolution ensemble encoding, 2) multi-resolution ensemble decoding, and 3) predictive coding.

\subsubsection{Multi-Resolution Ensemble Encoding}
The goal of the encoder in the proposed RAE-MEPC is to extract temporal features from an input time series at multiple scales. The encoder has $K^{(E)}$ sub-encoders with different encoding lengths. The $k$-th sub-encoder $E_k$ captures temporal behavior in time series of length $T^{(E_k)}$. This encoding length is defined as shown in Eq. (\ref{equ:equ1}).
%Equ 1
\begin{gather}
	\centering
	\begin{aligned}
		T^{(E_k)}=[\frac{1}{\tau^{k-1}} \times T], \quad 1 \leq k \leq K^{(E)} \label{equ:equ1}
	\end{aligned}
\end{gather}
where $\tau > 1$ is the hyperparameter determining the encoding length. Each sub-encoder receives a time series whose length matches its encoding length. To learn the multi-scale dependency of the original input time series, the input of each sub-encoder has a shape similar to that of the original input. Therefore, the input sequence of each sub-encoder $\mathrm{X}^{(E_k)}$ is obtained by downsampling the original time window $\mathrm{X}$ to a subsequence of length $T^{(E_k)}$, as shown in Eq. (\ref{equ:equ2}).
%Equ 2
\begin{gather}
	\centering
	\begin{aligned}		
		\mathrm{X}^{(E_k)}=[\mathrm{x}_1^{(E_k)}, \mathrm{x}_2^{(E_k)}, \dotsm \mathrm{x}_{T^{(E_k)}}^{(E_k)}]=\{\mathrm{x}_i\}, \\
		i = [1+j \times \frac{T-1}{T^{(E_k)}-1}], \quad 0 \leq j < T^{(E_k)}
		\label{equ:equ2}
	\end{aligned}
\end{gather}

%Algorithm1
\begin{algorithm*}[t!]
	\caption{Training of RAE-MEPC.}
	\begin{algorithmic}[1]
		\renewcommand{\algorithmicrequire}{\textbf{Input:}}
		\renewcommand{\algorithmicensure}{\textbf{Output:}}
		\REQUIRE a batch of time series $\{\textbf{X}_{b}\}$, a number of sub-encoders ${ K }^{ (E) }$, a number of reconstruction sub-decoders ${ K }^{ (RD) }$, resolution hyperparameter $\tau$, a number of batches $B$, a number of epochs $n$
		\ENSURE Encoder $E=\{{ E }_{1}, ..., { E }_{{ K }^{ (E) }} \}$, reconstruction decoder $RD=\{{ RD }_{1}, ..., { RD }_{{ K }^{ (RD) }} \}$
		
		\STATE Set encoding length ${ T }^{ ({E}_{k}) }=[\frac{1}{{\tau}^{k-1}} \times T]$ for each ${E}_{k}$
		\STATE Set decoding length ${ T }^{ ({D}_{k}) }=[\frac{1}{{\tau}^{k-1}} \times T]$ for each ${RD}_{k}$
		
		\FOR {$epoch=1, ..., n$}
		\FOR {$b=1, ..., B$}
		\FOR {$k=1, ..., { K }^{ (E) }$}
		\STATE Preprocess a input time series $\{\textbf{X}_{b}^{({E}_{k})}\}$ for ${E}_{k}$ via Eq. (\ref{equ:equ2})
		\STATE Feed $\{\textbf{X}_{b}^{({E}_{k})}\}$ to ${E}_{k}$ and obtain last hidden states $\{{\textbf{h} }_{ T^{({E}_{k})} }^{({E}_{k})}\}$ via Eq. (\ref{equ:equ3})
		\ENDFOR
		\STATE Obtain encoder representation ${\textbf{h} }^{ (E) }$ using hierarchical multi-resolution ensemble via Eq. (\ref{equ:equ4})-(\ref{equ:equ6})
		\FOR {$k={ K }^{ (RD) }, ..., 1$}
		\STATE Run the reconstruction sub-decoder ${RD}_{k}$
		\IF {$k\neq{ K }^{ (RD) }$}
		\STATE Perform coarse-to-fine fusion and obtain updated hidden states $\{{\textbf{\^{h} }_{ t }^{({RD}_{k})}}\}$ via Eq. (\ref{equ:equ11})
		\ENDIF
		\STATE Obtain hidden states $\{{\textbf{{h} }_{ t }^{({RD}_{k})}}\}$ and reconstruction outputs $\{{\textbf{{y}}_{ t }^{({RD}_{k})}}\}$
		\ENDFOR
		\STATE Feed ${\textbf{h} }^{ (E) }$ to prediction decoder $PD$ and obtain prediction outputs $\{{\textbf{{y}}_{ t }^{(PD)}}\}$
		\ENDFOR
		\STATE Calculate total loss via Eq. (\ref{equ:equ21}) and minimize it by Adam optimizer
		\ENDFOR
	\end{algorithmic}
	\label{algorithm1}
\end{algorithm*}

Fig. \ref{fig1} shows the example of the multi-resolution ensemble encoding in $K^{(E)}=3$, $T=12$, and $\tau=2$. Given an input subsequence of each sub-encoder, sequential data are compressed into a representation of a fixed length. In RAE-MEPC, all the sub-encoders have the LSTM architectures and achieve the features at different resolutions as shown in Eq. (\ref{equ:equ3}).
%Equ 3
\begin{gather}
	\centering
	\begin{aligned}		
		\mathrm{h}_t^{(E_k)}=\mathrm{LSTM}^{(E_k)}(\mathrm{x}_t^{(E_k)}; \mathrm{h}_{t-1}^{(E_k)}), \label{equ:equ3}
	\end{aligned}
\end{gather}
where $\mathrm{h}_t^{(E_k)}$ is the hidden state of $k$-th sub-encoder $E_k$ at $t$-th time step, while $\mathrm{x}_t^{(E_k)}$ is the input subsequence of $E_k$ at $t$-th time step. $\mathrm{LSTM}^{(E_k)}$ is the LSTM model of $E_k$, where the hidden state at the $t$-th time step is obtained from the previous hidden state $\mathrm{h}_{t-1}^{(E_k)}$ and the current input $\mathrm{x}_t^{(E_k)}$. In this process, all the sub-encoders independently outputs the last hidden state at different resolutions. The lower-resolution sub-encoder with shorter encoding length can extract macro temporal characteristics, whereas the higher-resolution sub-encoder with a longer encoding length can focus on local temporal patterns. Finally, the integrated encoded representation is obtained by hierarchically aggregating the last hidden states from the lowest resolution to the highest resolution, as follows:
%Equ 4
\begin{gather}
	\centering
	\begin{aligned}		
		\mathrm{h}^{(E_k)}=\mathrm{MLP}^{(E_k)}(\mathrm{h}_{T^{(E_k)}}^{(E_k)}), \quad k=K^{(E)}, \label{equ:equ4}
	\end{aligned}
\end{gather}
%Equ 5
\begin{gather}
	\centering
	\begin{aligned}		
		\mathrm{h}^{(E_k)}=\mathrm{MLP}^{(E_k-E_{k+1})}(\mathrm{h}_{T^{(E_k)}}^{(E_k)} + \mathrm{h}^{(E_{k+1})}), \space k \neq K^{(E)}, \label{equ:equ5}
	\end{aligned}
\end{gather}
%Equ 6
\begin{gather}
	\centering
	\begin{aligned}		
		\mathrm{h}^{(E)}=\mathrm{h}^{(E_1)}, \label{equ:equ6}
	\end{aligned}
\end{gather}
where $\mathrm{MLP}^{(E_k-E_{k+1})}$ is the fully connected layer that integrates the information of $E_k$ and $E_{k+1}$. $\mathrm{h}_{T^{(E_k)}}^{(E_k)}$ is the last hidden state of $E_k$ and $\mathrm{h}^{(E_k)}$ is the integrated features from $E_{K^{(E)}}$ to $E_k$. $\mathrm{h}^{(E)}$ is the final encoded representation obtained from the ensemble of all outputs of the sub-encoders with different resolutions. Finally, the encoder of RAE-MEPC can capture the multi-scale dependency with both general and local characteristics by introducing multi-resolution ensemble encoding.

\subsubsection{Multi-Resolution Ensemble Decoding}
The reconstruction decoder of RAE-MEPC aims to reconstruct input time series based on encoded representation effectively. It reconstructs the input time series in reverse order based on the multi-resolution ensemble decoding proposed in RAMED \cite{shen2021time}. The reconstruction decoder consists of $K^{(RD)}$ sub-decoders. Each sub-decoder has a different decoding length to encourage each sub-decoder to model the temporal patterns at different resolutions. The decoding length $T^{(RD_k)}$ of the $k$-th sub-decoder $RD_k$ is defined in the same manner as the encoding length as shown in Eq. (\ref{equ:equ7}).
%Equ 7
\begin{gather}
	\centering
	\begin{aligned}		
		T^{(RD_k)}=[\frac{1}{\tau^{k-1}} \times T], \quad 1 \leq k \leq K^{(E)} \label{equ:equ7}
	\end{aligned}
\end{gather}

The resolution of the sub-decoder is the same as that of the corresponding sub-encoder because they have the same length for encoding and decoding. In the reconstruction decoder, all the sub-decoders have an LSTM architecture. Each sub-decoder reconstructs the input time series of decoding length based on the final encoded representation without teacher forcing as follows:
%Equ 8
\begin{gather}
	\centering
	\begin{aligned}		
		\mathrm{y}_t^{(RD_k)}=\mathrm{h}^{(E)},\quad t=T^{(RD_k)}, \label{equ:equ8}
	\end{aligned}
\end{gather}
%Equ 9
\begin{gather}
	\centering
	\begin{aligned}		
		\mathrm{y}_t^{(RD_k)}=\mathrm{MLP}^{(RD_k)}(\mathrm{h}_t^{(RD_k)}),\quad t \neq T^{(RD_k)} \label{equ:equ9}
	\end{aligned}
\end{gather}
%Equ 10
\begin{gather}
	\centering
	\begin{aligned}		
		\mathrm{h}_{t-1}^{(RD_k)}=\mathrm{LSTM}^{(RD_k)}(\mathrm{y}_t^{(RD_k)}+\epsilon \delta; \hat{\mathrm{h}}_t^{(RD_k)}) \label{equ:equ10}
	\end{aligned}
\end{gather}
where $\mathrm{y}_t^{(RD_k)}$and $\mathrm{h}_{t}^{(RD_k)}$ are the output and the hidden states of the $k$-th sub-decoder $RD_k$ at $t$-th time step, respectively. $\mathrm{MLP}^{(RD_k)}$ is the fully connected layer for the output, while $\mathrm{LSTM}^{(RD_k)}$ is the LSTM model to capture temporal dependency in $RD_k$. For each sub-decoder, a small amount of noise $\epsilon \delta$ is added to the input of LSTM. It is for improving robustness to partial corruption of the input pattern as in the denoising auto-encoder \cite{vincent2008extracting}. In RAMED \cite{shen2021time}, the reconstruction decoder can mitigate the error accumulation of the RAE using multi-resolution ensemble decoding. For this decoding, RAMED utilizes a coarse-to-fine fusion strategy to fuse a lower-resolution sub-decoder with a higher-resolution sub-decoder. It combines the hidden state of $RD_k$ with the information extracted from the coarser-grained sub-decoder $RD_{k+1}$. From this strategy, the integrated hidden state $\hat{\mathrm{h}}_t^{(RD_k)}$ is obtained as shown in Eq. (\ref{equ:equ11}).
%Equ 11
\begin{gather}
	\centering
	\begin{split}
		\hat{\mathrm{h}}_t^{(RD_k)}= {}& \beta \mathrm{h}_{t+1}^{(RD_k)}\\
		&+(1-\beta)\mathrm{MLP}^{(RD_k-RD_{k+1})} ([\mathrm{h}_{t+1}^{(RD_k)}; \mathrm{h}_{[t/\tau]}^{(RD_{k+1})}]),
		\label{equ:equ11}
	\end{split}
\end{gather}
where $\mathrm{h}_{t+1}^{(RD_k)}$ is the previous hidden state of $RD_k$, while $\mathrm{h}_{[t/\tau]}^{(RD_{k+1})}$ is the corresponding hidden state of a nearby coarser sub-decoder $RD_{k+1}$. $\beta$ is the hyperparameter adjusting the degree to reflect the information in coarser-grained decoders. More lower-resolution information was exploited in each decoding step with a smaller $\beta$. From this process, the reconstruction decoder can use multi-resolution information by directly using the coarser-grained information to decode of the finer-grained decoder. After multi-resolution ensemble decoding, the reconstructed input time series in reverse order can be obtained from the highest resolution decoder. Finally, we achieve the final reconstructed output $\overleftarrow{\mathrm{Y}}_{recon}$ by reversing the output of reconstruction decoder as shown in Eq. (\ref{equ:equ12}).
%Equ 12
\begin{gather}
	\centering
	\begin{aligned}		
		\overleftarrow{\mathrm{Y}}_{recon} = [\mathrm{y}_1^{(RD_1)}, \mathrm{y}_2^{(RD_1)}, \dotsm, \mathrm{y}_T^{(RD_1)}],
		\label{equ:equ12}
	\end{aligned}
\end{gather}

\subsubsection{Predictive Coding}
We introduce predictive coding to extract the long-term dependency of input time series. The goal of the prediction decoder is to predict the future time series after $T/2$ time steps based on compressed representation from the encoder. In the proposed model, the prediction decoder is LSTM architecture to perform multi-step prediction tasks simply as follows.
%Equ 13
\begin{gather}
	\centering
	\begin{aligned}		
		\mathrm{h}_0^{(PD)}=\mathrm{h}^{(E)},
		\label{equ:equ13}
	\end{aligned}
\end{gather}
%Equ 14
\begin{gather}
	\centering
	\begin{aligned}		
		\mathrm{h}_t^{(PD)}=\mathrm{LSTM}^{(PD)}(\mathrm{x}_t; \mathrm{h}_{t-1}^{(PD)}),
		\label{equ:equ14}
	\end{aligned}
\end{gather}
%Equ 15
\begin{gather}
	\centering
	\begin{aligned}		
		\mathrm{y}_t^{(PD)}=\mathrm{MLP}^{(PD)}(\mathrm{h}_t^{(PD)}),
		\label{equ:equ15}
	\end{aligned}
\end{gather}
where $\mathrm{h}_t^{(PD)}$ and $\mathrm{y}_t^{(PD)}$ are the hidden state and the predicted output at $t$-th time step in the prediction decoder, respectively. $\mathrm{MLP}^{(PD)}$ is the fully connected layer for the output, whereas $\mathrm{LSTM}^{(PD)}$ is the LSTM model in the prediction decoder used to model temporal dependency. Finally, the prediction decoder predicts the future time series $\overrightarrow{\mathrm{Y}}_{pred}$ after $T/2$ time step as shown in Eq. (\ref{equ:equ16}). The prediction decoder enables the encoder to learn the more informative patterns in normal time series by reflecting the temporal dependency from the prediction task's perspective.
%Equ 16
\begin{gather}
	\centering
	\begin{aligned}		
		\overrightarrow{\mathrm{Y}}_{pred} = [\mathrm{y}_1^{(PD)}, \mathrm{y}_2^{(PD)}, \dotsm, \mathrm{y}_T^{(PD)}],
		\label{equ:equ16}
	\end{aligned}
\end{gather}

%Algorithm2
\begin{algorithm}[t!]
	\caption{Anomaly Detection of RAE-MEPC.}
	\begin{algorithmic}[1]
		\renewcommand{\algorithmicrequire}{\textbf{Input:}}
		\renewcommand{\algorithmicensure}{\textbf{Output:}}
		\REQUIRE a validation set ${\textbf{X}}^{valid}=\{{ \textbf{X} }_{ 1 }^{valid}, ..., { \textbf{X} }_{ N_{valid} }^{valid}\}$, a test set $\textbf{X}^{test}=\{{ \textbf{X} }_{ 1 }^{test}, ..., { \textbf{X} }_{ N_{test} }^{test}\}$, Encoder $E$, reconstruction decoder $RD$, predefined threshold $THR$
		\ENSURE Anomaly labels of all time series in the test set
		
		\FOR {$i=1, ..., N_{valid}$}
    		\STATE Feed ${ \textbf{X} }_{ i }^{valid}$ to $E$ and obtain ${\textbf{h} }^{ (E) }$
    		\STATE Feed ${\textbf{h} }^{ (E) }$ to $RD$ and obtain $\{\mathrm{y}_t^{(RD_1)}\}$
    		\STATE Calculate residual $\mathrm{e}_t^{valid}$ for each time step via Eq. (\ref{equ:equ22})
		\ENDFOR
		\STATE Estimate $\mathcal{N}(\mu, \Sigma)$ of $\{\mathrm{e}_t^{valid}\}$
		\FOR {$i=1, ..., N_{test}$}
    		\STATE Feed ${ \textbf{X} }_{ i }^{test}$ to $E$ and obtain ${\textbf{h} }^{ (E) }$
    		\STATE Feed ${\textbf{h} }^{ (E) }$ to $RD$ and obtain $\{\mathrm{y}_t^{(RD_1)}\}$
    		\STATE Calculate residual $\mathrm{e}_t^{test}$ and obtain $Anomaly \: Score$ for each time step via Eq. (\ref{equ:equ22})-(\ref{equ:equ23})
    		\IF {$Anomaly \: Score > THR$}
    		    \STATE Anomaly label of $\mathrm{x}_t^{test}$ is abnormal
    		\ELSE
    		    \STATE Anomaly label of $\mathrm{x}_t^{test}$ is normal
    		\ENDIF
		\ENDFOR
	\end{algorithmic}
	\label{algorithm2}
\end{algorithm}

%Table 1
\begin{table*}[!t]
    \renewcommand{\arraystretch}{1.3}
    \caption{Dataset description.}
    \label{table1}
    \centering
    \resizebox{\textwidth}{!}{
        \begin{tabular}{ccccccc} 
        	\hline
        	Dataset      & \begin{tabular}[c]{@{}c@{}}\# Variables\end{tabular} & \begin{tabular}[c]{@{}c@{}}Length of \\Time Window\end{tabular} & \begin{tabular}[c]{@{}c@{}}Stride of \\Sliding Window\end{tabular} & \begin{tabular}[c]{@{}c@{}}\# Training\end{tabular} & \begin{tabular}[c]{@{}c@{}}\# Validation\end{tabular} & \begin{tabular}[c]{@{}c@{}}\# Test\end{tabular}  \\ 
        	\hline
        	Power-demand & 1                                                      & 512                                                             & 256                                                                & 12,701                                                     & 5,444                                                          & 14,786                                                     \\ 
        	\hline
        	2D-gesture   & 2                                                      & 64                                                              & 32                                                                 & 5,775                                                    & 2,476                                                          & 3,000                                                     \\
        	\hline
        \end{tabular}}
\end{table*}

\subsection{Objective Function}
The total loss of RAE-MEPC has three loss terms as follows: 1) reconstruction error $L_{recon}$, 2) multi-resolution shape-forcing loss $L_{shape}$, and 3) prediction error $L_{pred}$. First, $L_{recon}$ is the mean squared error for the difference between the input time series and the reconstructed output as shown in Eq. (\ref{equ:equ18}). This loss can encourage the reconstructed output to be close to the input time series.
%Equ 17
\begin{gather}
	\centering
	\begin{aligned}		
		L_{recon} = \sum_{t=1}^{T}\lVert \mathrm{y}_t^{(RD_1)} - \mathrm{x}_t \lVert^2_2
		\label{equ:equ17}
	\end{aligned}
\end{gather}

Second, $L_{shape}$ is the multi-resolution shape-forcing loss proposed in RAMED. This loss is introduced to encourage the sub-decoders in the reconstruction decoder to learn consistent temporal trends as the original input time series. It can force sub-decoders at different resolutions to learn similar temporal patterns as the input in multi-resolution ensemble decoding. Because $L_{recon}$ already makes the reconstructed output similar to the input time series, $L_{shape}$ is defined in the sub-decoders except for the sub-decoder with the highest resolution. This loss is based on DTW \cite{sakoe1978dynamic}, which can calculate the distance between two time series data with different lengths. However, the DTW distance is non-differentiable because it contains a min operator. Therefore, the multi-resolution shape-forcing loss introduces smoothed DTW (sDTW) \cite{cuturi2017soft}, which computes the soft minimum of all alignment costs in the DTW distance. Eq. (\ref{equ:equ19}) shows the sDTW distance where the smoothed minimum operator is introduced. Finally, $L_{shape}$ is based on the sDTW distance between the input time series $\mathrm{X}$ and the output of the $k$-th sub-decoder $\overleftarrow{\mathrm{Y}}^{(RD_k)}$ as shown in Eq. (\ref{equ:equ18}).
%Equ 18
\begin{gather}
	\centering
	\begin{aligned}		
		sDTW(\mathrm{X}, \overleftarrow{\mathrm{Y}}^{(RD_k)}) = -\gamma \log{\sum_{\mathrm{A}\in\mathcal{A}}e^{-\langle\mathrm{A} \mathrm{C}\rangle/\gamma}},
		\label{equ:equ18}
	\end{aligned}
\end{gather}
%Equ 19
\begin{gather}
	\centering
	\begin{aligned}		
		L_{shape} = \frac{1}{K^{(RD)}-1}\sum_{k=2}^{K^{(RD)}} sDTW(\mathrm{X}, \overleftarrow{\mathrm{Y}}^{(RD_k)}),
		\label{equ:equ19}
	\end{aligned}
\end{gather}
where $\mathrm{A}$ is the alignment matrix of the DTW alignment path, $\mathrm{C}$ is the matrix for alignment costs, whose element is the Euclidean distance between the corresponding elements of the two time series, and $\gamma$ is a hyperparameter related to the smoothed minimum operator. More detailed information on the multi-resolution shape-forcing loss can be found in \cite{shen2021time}.

Finally, $L_{pred}$ is the square of the difference between the predicted output and the actual future time series as shown in Eq. (\ref{equ:equ20}). It encourages the prediction decoder to predict the future time series after the $T/2$ time step effectively. It also enables the encoder to capture additional temporal information, which is helpful for the prediction task.
%Equ 20
\begin{gather}
	\centering
	\begin{aligned}		
		L_{pred} = \sum_{t=1}^{T}\lVert \mathrm{y}_t^{(PD)} - \mathrm{x}_{t+[T/2]} \lVert^2_2
		\label{equ:equ20}
	\end{aligned}
\end{gather}

From above three loss terms, RAE-MEPC is trained by minimizing the total loss $L_{total}$ as shown in Eq. (\ref{equ:equ21}). The weights $\lambda_{shape}$ and $\lambda_{pred}$ represent the hyperparameters that control the importance of $L_{shape}$ and $L_{pred}$, respectively. The detailed training procedure of the proposed RAE-MEPC is shown in Algorithm \ref{algorithm1}.
%Equ 21
\begin{gather}
	\centering
	\begin{aligned}		
		L_{total} = L_{recon} + \lambda_{shape}L_{shape} + \lambda_{pred}L_{pred}
		\label{equ:equ21}
	\end{aligned}
\end{gather}

\subsection{Anomaly Detection}
Once the RAE-MEPC is trained by minimizing the objective function, anomalies can be detected using the reconstruction error. Specifically, the anomalies in the test set can be detected based on the distribution of the reconstruction error in the validation set. Given a time series in the validation set, the residual $\mathrm{e}_t$ is obtained at each time step, as shown in Eq. (\ref{equ:equ22}). Then we estimate the distribution $\mathrm{e}_t \sim \mathcal{N}(\mu, \Sigma)$ of the normal residuals based on the maximum likelihood estimation. Based on this distribution, we can detect anomalies in the unseen time series in the test set. The anomaly score is calculated using Eq. (\ref{equ:equ23}). This score represents the degree to which the residual of the given test data deviates from the estimated normal distribution. Finally, the instances are detected as anomalies if the anomaly score exceeds the predefined threshold. The detailed anomaly detection procedure of the proposed RAE-MEPC is shown in Algorithm \ref{algorithm2}.
%Equ 22
\begin{gather}
	\centering
	\begin{aligned}		
		\mathrm{e}_t=\mathrm{y}_t^{(RD_1)}-\mathrm{x}_t
		\label{equ:equ22}
	\end{aligned}
\end{gather}
%Equ 23
\begin{gather}
	\centering
	\begin{aligned}		
		Anomaly \: Score= (\mathrm{e}_t - \mu)^T \Sigma^{-1}(\mathrm{e}_t - \mu)
		\label{equ:equ23}
	\end{aligned}
\end{gather}

%Table 2
\begin{table*}[!t]
	\renewcommand{\arraystretch}{1.3}
	\caption{Overall performance comparison.}
	\label{table2}
	\centering
    \resizebox{\textwidth}{!}{
    	\begin{tabular}{c|c|ccc|ccc} 
    		\hline
    		Model Type   & Model                                  & \multicolumn{3}{c|}{2D-gesture}                                             & \multicolumn{3}{c}{Power-demand}                                            \\ 
    		\cline{3-8}
    		&                                                         & Best F1-Score           & AUROC                   & AUPRC                   & Best F1-Score           & AUROC                   & AUPRC                    \\ 
    		\hline
    		Prediction                      & LSTM-AD                                                 & 0.5255                  & 0.7409                  & 0.5081                  & 0.2549                  & 0.6580                  & 0.1704                   \\ 
    		\hline
    		Reconstruction (GAN)                      & MAD-GAN                                                 & 0.4057                  & 0.4867                  & 0.2323                  & 0.2621                  & 0.6081                  & 0.1235                   \\ 
    		\hline
    		Reconstruction & EncDec-AD                                               & 0.5573                  & 0.7745                  & 0.5457                  & 0.2103                  & 0.5034                  & 0.0982                   \\ 
    		\cline{2-8}
    		& RAMED & 0.5625                  & 0.7813                  & 0.5797                  & \textbf{\underline{0.2953}} & 0.6905                  & 0.1787                   \\ 
    		\cline{2-8}
    		& \begin{tabular}[c]{@{}c@{}} RAE-MEPC \\ (ours)\end{tabular}                                                    & \textbf{\underline{0.5685}} & \textbf{\underline{0.7973}} & \textbf{\underline{0.5915}} & 0.2940                  & \textbf{\underline{0.6931}} & \textbf{\underline{0.2372}}  \\
    		\hline
    	\end{tabular}}
\end{table*}
%Table 3
\begin{table*}[!t]
	\renewcommand{\arraystretch}{1.1}
	\caption{Effects of multi-resolution ensemble encoding and predictive coding.}
	\label{table3}
	\centering
	\resizebox{\textwidth}{!}{
    	\begin{tabular}{c|ccc|ccc} 
    		\hline
    		Model                  & \multicolumn{3}{c|}{2D-gesture}                                                                     & \multicolumn{3}{c}{Power-demand}                                                            \\ 
    		\cline{2-7}
    		& Best F1-Score                   & AUROC                           & AUPRC                           & Best F1-Score           & AUROC                           & AUPRC                            \\ 
    		\hline
    		\begin{tabular}[c]{@{}c@{}} w/o multi-resolution \\ ensemble encoding \end{tabular}& 0.5681                          & 0.7898                          & 0.5575                          & 0.2794                  & 0.6871                          & 0.1786                           \\ 
    		\hline
    		w/o predictive coding & 0.5683                          & 0.7952                          & 0.5858                          & 0.2886                  & \textbf{\underline{0.6953}} & 0.1857                           \\ 
    		\hline
    		\hline
    		Full model                                    & \textbf{\underline{0.5685}} & \textbf{\underline{0.7973}} & \textbf{\underline{0.5915}} & \textbf{\underline{0.2940}} & 0.6931 & \textbf{\underline{0.2372}}  \\
    		\hline
    	\end{tabular}}
\end{table*}

\section{Experiments}
\subsection{Experimental Settings}
\subsubsection{Data}
In this study, we evaluate our proposed method on the two real-world benchmark datasets for time-series anomaly detection: power-demand\footnote{\url{http://www.cs.ucr.edu/~eamonn/discords/power_data.txt}}\cite{keogh2005hot} and 2D-gesture\footnote{\url{http://www.cs.ucr.edu/~eamonn/discords/ann_gun_CentroidA}}\cite{keogh2005hot}.

\textbf{\textit{Power-demand dataset.}} This dataset contains the power consumption measured in a Dutch research facility for the entire year of 1997. In this dataset, normal patterns of power demand from 9 am to 5 pm on an ordinary week (Monday to Friday) were accumulated. Anomalies were measured during unusual weeks such as holidays.

\textbf{\textit{2D-gesture dataset.}} This dataset consists of the X and Y coordinates of the right hand of actors. These coordinates were extracted from video images where the actor grabs a gun from a hip-mounted holster, moves it to the target, and returns it to the holster. The anomalies were defined as the scenes where actors did not return the gun to the holster.

We followed the experimental setting in \cite{shen2021time}. For power-demand and 2D-gesture, the raw dataset has only a training set and a test set. We used 30\% of the training set as the validation set in each dataset to allow model selection and hyperparameter tuning. The raw training data are partitioned into a time window of fixed length using a sliding window to design the temporal data. In this study, the length of the time window was set as 512 for the power-demand dataset and 64 for the 2D-gesture dataset. We set the sliding window to have a stride of 256 on the power-demand dataset and 32 for the 2D-gesture dataset. For a fair comparison, we employed the same data pre-processing for all methods. Table \ref{table1} summarizes the dataset statistics.

\subsubsection{Baseline Methods}
The proposed model was compared to the following three categories of multivariate time-series anomaly detection algorithms. The first category is the prediction-based anomaly detection model: LSTM-ED \cite{malhotra2015long}. Its architecture is a stacked LSTM to detect anomalies in a time series. This network was trained on only normal data and was used as a predictor over several time steps. The prediction errors were used as anomaly scores by modeling them as a multivariate Gaussian distribution. The second category is the reconstruction-based GAN method, and MAD-GAN \cite{li2019mad} is employed as a representative model for this category. It is an unsupervised multivariate anomaly detection method based on a GAN consisting of a generator and discriminator. In the MAD-GAN, a novel anomaly score called the DR-score was used to detect anomalies based on discrimination and reconstruction. The third category is reconstruction-based methods, in which our proposed method belongs. EncDec-AD \cite{malhotra2016lstm} and RAMED \cite{shen2021time} were employed as benchmark models. EncDec-AD was trained to reconstruct normal time-series behavior based on the RAE architecture. In this model, the reconstruction errors were used to detect anomalies. RAMED is a simple yet efficient recurrent network ensemble. By using decoders with different decoding lengths and a new coarse-to-fine fusion mechanism, lower-resolution information can aid in long-range decoding for decoders with higher-resolution outputs. In this model, we used the output from the decoder with the highest resolution to obtain an anomaly score at each time step.

\subsubsection{Evaluation Metrics}
We evaluated the proposed RAE-MEPC using the following metrics for time-series anomaly detection: area under the ROC curve (AUROC), area under the precision-recall curve (AUPRC), and the best F1-score. AUROC and AUPRC were used to evaluate the threshold-independent intrinsic anomaly detection ability of the model. AUROC measures the entire two-dimensional area under the ROC curve, which measures how accurately the model can detect true anomalies under a certain level of false alarm. This score indicates the performance of the anomaly detector at various threshold settings. AUPRC shows the trade-off between precision and recall for different thresholds. A high score represents both high recall and high precision, where high precision relates to a low false-positive rate, and high recall relates to a low false-negative rate. The best F1-score is the highest F1-score at the different thresholds, and it is selected from the F1-scores in 1,000 thresholds uniformly distributed from zero to the maximum anomaly score in the test dataset.

%Figure 2
\begin{figure*}[!t]
    \centering
    \subfloat[Input time series.\label{fig2a}]{%
       \includegraphics[width=0.85\linewidth]{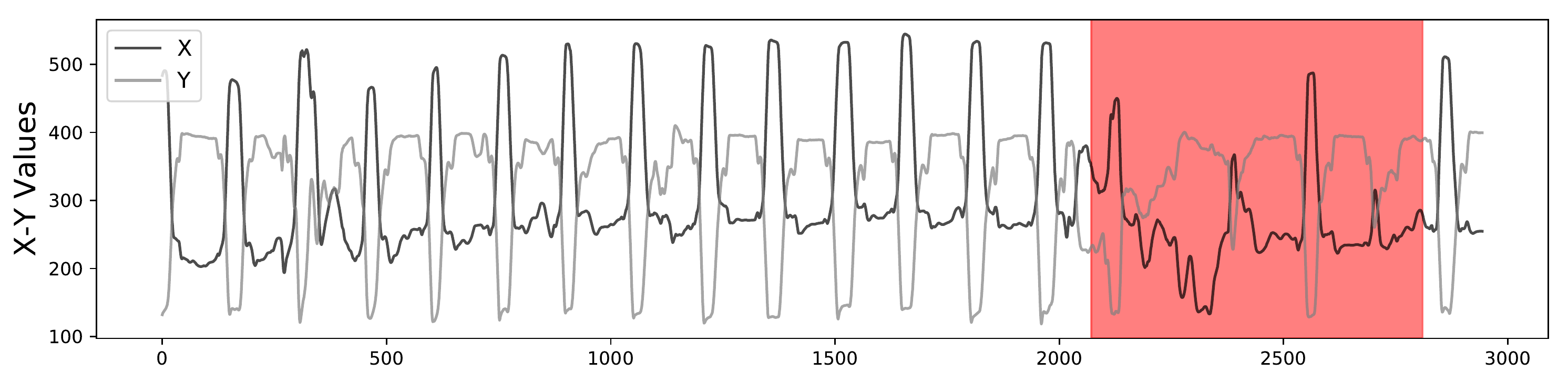}} \\
    \subfloat[Reconstructed output of RAE-MEPC.\label{fig2b}]{%
        \includegraphics[width=0.85\linewidth]{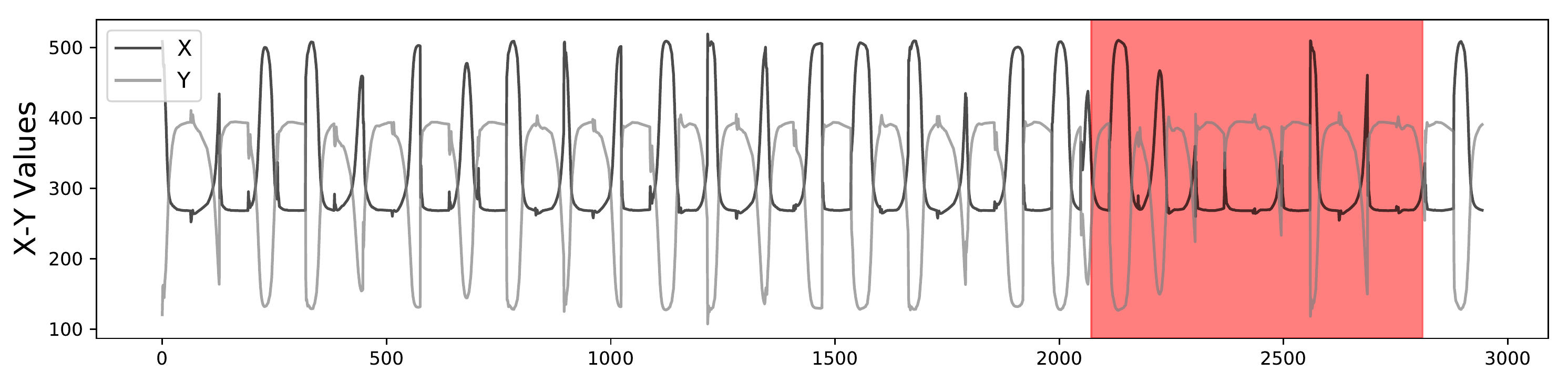}} \\
    \subfloat[Anomaly score.\label{fig2c}]{%
        \includegraphics[width=0.85\linewidth]{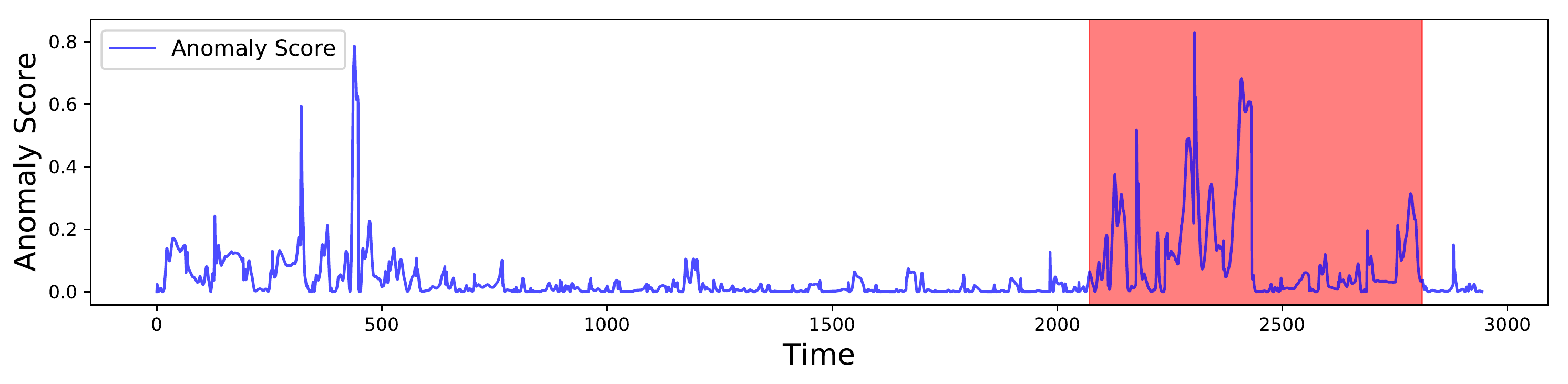}}
    \caption{Comparison of input time series and reconstructed output from RAE-MEPC.}
    \label{fig2}
\end{figure*}

\subsubsection{Implementation Details}
In this study, we used the encoder with three sub-encoders at different resolutions. We also built the reconstruction decoder using three sub-decoders with different decoding length. In RAE-MEPC, sub-encoder, sub-decoder, and prediction decoder were based on the single-layer LSTM architectures. We performed grid search on the hyperparameters as follows: encoder/decoder hidden dimension in $\{16, 32, 64\}$, $\tau$ in $\{2, 3, 4\}$, $\beta$ in $\{0.1, 0.3\}$, and $\lambda_{shape}$ in $\{0.0001, 0.001\}$. We set $\lambda_{pred}$ to one. We trained our proposed model using Adam optimizer \cite{kingma2014adam} with an initial learning rate of 0.001. All experiments were performed on a Linux workstation with Intel Core i7-9700X CPU, 128 GB RAM, and NVidia GeForce RTX 3080 GPU using PyTorch.

%Table 4
\begin{table*}[!t]
	\renewcommand{\arraystretch}{1.2}
	\caption{Effect of $\tau$.}
	\label{table4}
	\centering
	\resizebox{0.9\textwidth}{!}{
    	\begin{tabular}{c|ccc|ccc} 
    		\hline
    		$\quad\quad\quad\tau\quad\quad\quad$ & \multicolumn{3}{c|}{2D-gesture}                                                                     & \multicolumn{3}{c}{Power-demand}                                                    \\ 
    		\cline{2-7}
    		& Best F1-Score                   & AUROC                           & AUPRC                           & Best F1-Score           & AUROC                   & AUPRC                            \\ 
    		\hline
    		2                                    & 0.5463                          & 0.7574                          & 0.5152                          & 0.2681                  & 0.6145         & 0.1279                           \\ 
    		\hline
    		3                                    & 0.5522                          & 0.7722                          & 0.5640                          & 0.2473                  & 0.6303         & 0.1511                           \\ 
    		\hline
    		4                                    & \textbf{\underline{0.5685}} & \textbf{\underline{0.7973}} & \textbf{\underline{0.5915}} & \textbf{\underline{0.2940}} & \textbf{\underline{0.6931}} & \textbf{\underline{0.2372}}  \\
    		\hline
    	\end{tabular}}
\end{table*}
%Figure 3
\begin{figure*}[!t]
    \centering
    \subfloat[2D-gesture dataset.\label{fig3a}]{%
       \includegraphics[width=0.45\linewidth]{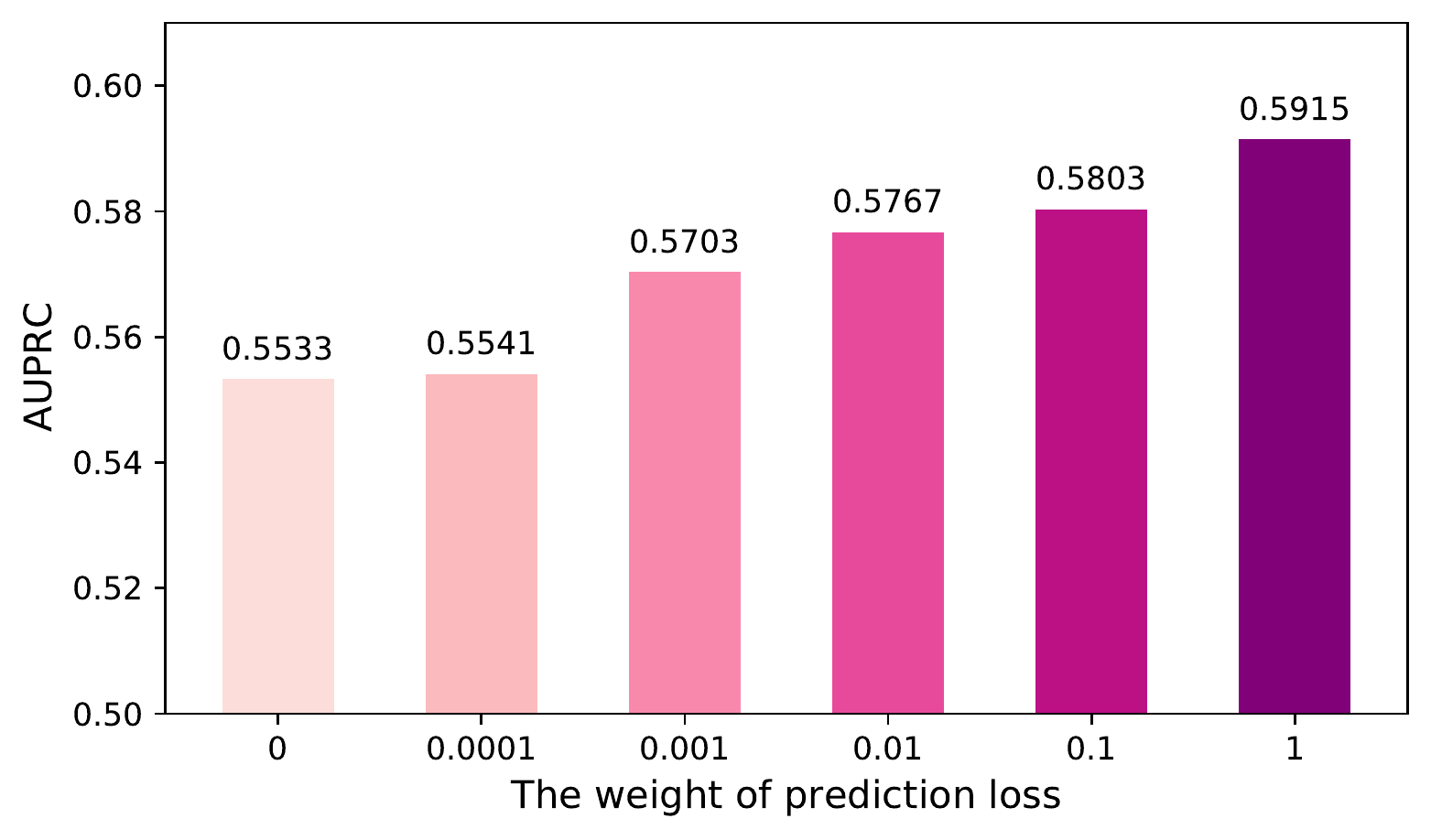}}
    \hfil
    \subfloat[Power-demand dataset.\label{fig3b}]{%
        \includegraphics[width=0.45\linewidth]{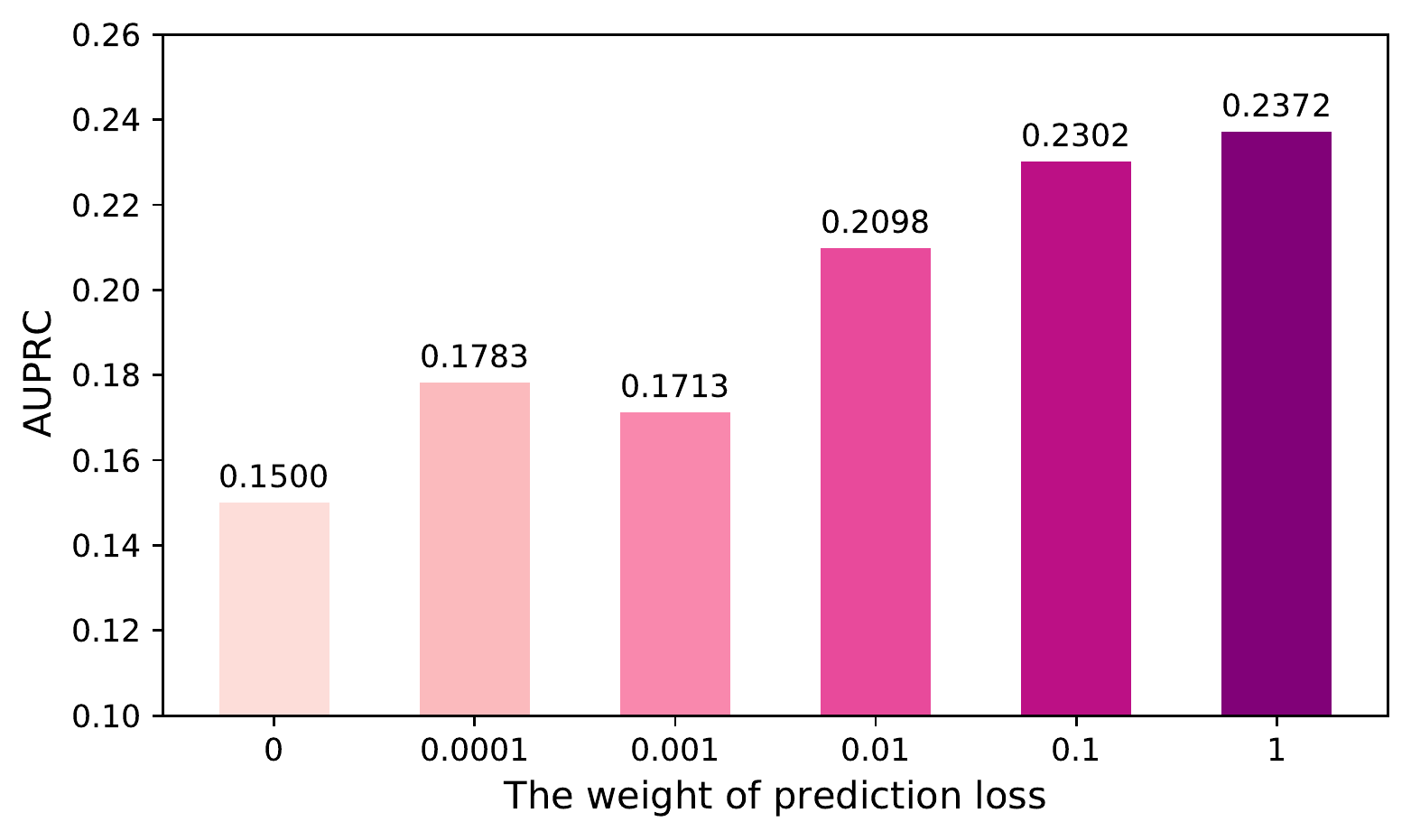}}
    \caption{Effect of varying $\lambda_{pred}$.}
    \label{fig3}
\end{figure*}

\subsection{Experimental Results}
\subsubsection{Overall Performance Comparison}
Our proposed method RAE-MEPC was compared with existing prediction-based and reconstruction-based methods. A hyperparameter search was performed for all models. Table \ref{table2} presents the best results of the proposed model and baseline methods. RAE-MEPC outperformed the baseline models for all three metrics on the 2D-gesture dataset. It also showed the best performance on AUROC and AUPRC for the power-demand dataset. In other words, our proposed method achieved a higher performance than prediction-based method and reconstruction-based method with GAN showing the most outstanding performance among reconstruction-based methods where it belongs. These results demonstrate that using multi-resolution ensemble encoding and predictive coding has a positive impact on learning rich normal time-series representations, and helps the model detect anomalies effectively. From these results, we can also conclude that RAE-MEPC works well on both the univariate and multivariate datasets. However, MAD-GAN achieved lower performance on the 2D-gesture dataset compared to the power-demand dataset. This showed that GAN-based models usually have training difficulties because they can easily suffer from mode collapse and non-convergence problems. Furthermore, among the three metrics, the improvement in AUPRC is particularly significant. The proposed method achieved a low false positive rate and low false negative rate compared to other methods.

As shown in Fig. \ref{fig2}, we qualitatively evaluated the proposed reconstruction-based method. Fig. \ref{fig2a} shows the raw input time series of the test set on the 2D-gesture while Fig. \ref{fig2b} presents the reconstructed output of RAE-MEPC. The black and gray lines indicate the X- and Y-coordinates, respectively. The red region represents the time steps when the true anomalies were obtained. The reconstructed time series showed similar trends with the input time series on two variables in a normal period. However, RAE-MEPC also reconstructed the samples that followed an estimated normal distribution during an abnormal period. Therefore, high anomaly scores were derived based on the significant difference between the actual and reconstructed values in this abnormal region, as shown in Fig. \ref{fig2c}. From these results, we can conclude that the proposed RAE-MEPC learns rich normal time-series representations by introducing multi-resolution ensemble encoding and predictive coding.

\subsubsection{Effects of Model Components}
In this experiment, we evaluated the effects of multi-resolution ensemble encoding and predictive coding in the proposed model RAE-MEPC. We considered these model components by eliminating one component from the full model. We used the encoder with three sparsely connected RNN-based sub-encoders for the model without multi-resolution ensemble encoding, similar to RAMED. Without predictive coding, the model only performed multi-resolution ensemble reconstruction, and it was trained only on the sum of reconstruction and multi-resolution shape-forcing losses. Table \ref{table3} contains the best performances obtained in hyperparameter search for all models. As shown in Table \ref{table3}, both multi-resolution ensemble encoding and predictive coding are essential. Without multi-resolution ensemble encoding, the performance decreased for all datasets and evaluation metrics. This result demonstrated that it is helpful to merge the features extracted from input time series on different resolutions. The model resulted in lower performance on all datasets and evaluation metrics without predictive coding. From this result, we can conclude that extracting temporal features of normal data from the perspective of a prediction task can help achieve rich time-series representation.

\subsubsection{Sensitivity to Hyperparameters}
We conducted a sensitivity analysis for the following two hyperparameters: 1) resolution hyperparameter $\tau$ in Eq. (\ref{equ:equ1}) and Eq. (\ref{equ:equ7}), and 2) weight of prediction loss $\lambda_{pred}$ in Eq. (\ref{equ:equ21}). We set the default hyperparameters to $\beta=0.1$, $\tau=4$, $\lambda_{pred}=1$, and encoder and decoder hidden dimension of 32. The $\lambda_{shape}$ was set to 0.001 for 2D-gesture and 0.0001 for power-demand dataset. Table \ref{table4} shows the performances with different $\tau$ from two to four. When $\tau$ is four, RAE-MEPC exhibited the best performance on all evaluation metrics. As shown in Table \ref{table4}, increasing $\tau$ can improve performance as more diverse multi-resolution temporal patterns are handled in the encoder and decoder. Therefore, these experimental results demonstrated the effects of multi-resolution ensemble in encoding and decoding steps.

Finally, we studied the effects of prediction loss by increasing $\lambda_{pred}$ from zero to one on the two datasets. Fig. \ref{fig3} shows AUPRC with different $\lambda_{pred}$. RAE-MEPC achieved the highest AUPRC when $\lambda_{pred}$ was set to one. As shown in Fig. \ref{fig3}, when $\lambda_{pred}$ was large, better performance was achieved on all datasets. From these results, we can notice that it is helpful in time-series anomaly detection to learn the normal features from the perspective of a prediction task as well as a reconstruction task. 

\section{Conclusion}
Time-series anomaly detection has been employed in industries where large-scale time series can be accessed easily, such as smart factories. Time-series anomaly detection can improve productivity and reduce economic loss by monitoring potential risks and preventing faults of the systems. Therefore, it is crucial to detect anomalies accurately in the real world. Extracting rich temporal features from normal data can improve time-series anomaly detection performance.

In this paper, we proposed RAE-MEPC for unsupervised multivariate time-series anomaly detection. RAE-MEPC improves the quality of time-series representation using multi-resolution ensemble and predictive coding. This study introduced multi-resolution ensemble encoding to learn multi-scale dependency from input time series. The encoder achieves rich time-series representation by hierarchically combining the features extracted from sub-encoders with different resolutions. The proposed model reconstructs input time series using multi-resolution ensemble decoding on a coarse-to-fine method from this encoded representation. Moreover, we introduced predictive coding to extract temporal features from the perspective of a prediction task, and it encourages the encoder to capture more temporal features. Experiments on various time-series benchmark datasets demonstrated that the proposed model can detect anomalies more accurately than the well-known benchmark models.

Despite the favorable anomaly detection performance of the proposed RAE-MEPC, there are some limitations in the current work, which lead us to future research directions. First, although RAE-MEPC is effective in capturing multi-resolution temporal information of time-series data, it is weak in modeling the inter-correlation between variables. In future work, we will focus on an anomaly detection approach that can simultaneously model the inter-metric and temporal dependency for multivariate time series. Second, the proposed method causes some false alarms, although it significantly aids in detecting anomalies. In future research, we will improve the post-processing steps to reduce the number of false positives.

\bibliographystyle{abbrv}
\bibliography{refs}
\end{document}